\documentclass[letterpaper]{article} 
\usepackage{aaai2026}  
\usepackage{algorithmic}
\usepackage{algorithm}
\usepackage{amsmath}
\usepackage{booktabs}
\usepackage{amssymb}
\usepackage{bm}
\usepackage{arydshln}
\usepackage{xcolor}
\usepackage{graphicx}
\usepackage{times}  
\usepackage{helvet}  
\usepackage{courier}  
\usepackage[hyphens]{url}  
\usepackage{graphicx} 
\usepackage{natbib}  
\usepackage{caption} 
\usepackage{algorithm}
\usepackage{algorithmic}

\usepackage{newfloat}
\usepackage{listings}
\DeclareCaptionStyle{ruled}{labelfont=normalfont,labelsep=colon,strut=off} 
\floatstyle{ruled}
\newfloat{listing}{tb}{lst}{}
\floatname{listing}{Listing}
\title{DiTVR: Zero-Shot Diffusion Transformer for Video Restoration}
\author{
    Sicheng Gao,
    Nancy Mehta,
    Zongwei Wu,
    Radu Timofte
}
\affiliations{
    Computer Vision Lab, CAIDAS \& IFI, University of Würzburg\\
    Würzburg, Germany
}

\usepackage{bibentry}

\begin{document}

\maketitle

\begin{abstract}

Video restoration aims to reconstruct high-quality video sequences from low-quality inputs, addressing tasks such as super-resolution, denoising, and deblurring. Traditional regression-based methods often produce unrealistic details and require extensive paired datasets, while recent generative diffusion models face challenges in ensuring temporal consistency. We introduce \emph{DiTVR}, a zero‑shot video restoration framework that couples a diffusion transformer with {trajectory‑aware attention} and a {wavelet‑guided, flow‑consistent sampler}. Unlike prior 3D convolutional or frame‑wise diffusion approaches, our attention mechanism aligns tokens along optical‑flow trajectories, with particular emphasis on vital layers that exhibit the highest sensitivity to temporal dynamics. A {spatiotemporal neighbour cache} dynamically selects relevant tokens based on motion correspondences across frames. The flow-guided sampler injects data consistency only into low‑frequency bands, preserving high‑frequency priors while accelerating convergence. DiTVR establishes a new zero‑shot state of the art on video restoration benchmarks, demonstrating superior temporal consistency and detail preservation while remaining robust to flow noise and occlusions. 
\end{abstract}


\section{Introduction}
\label{sec:intro}

Video restoration seeks to recover high-quality ({HQ}) sequences from degraded ({LQ}) inputs, spanning super-resolution~\citep{kappeler2016video,rota2023enhancing}, denoising~\citep{li2023simple,ji2023spatio}, and deblurring~\citep{zhang2018adversarial,zhou2019spatio}.  Current state-of-the-art methods fall into two families: \emph{regression-based} approaches~\citep{caballero2017real,huang2017video,chan2021basicvsr} and \emph{generative} (diffusion-driven) approaches~\citep{lucas2019generative,yi2020progressive,xu2024videogigagan}.

Regression-based methods, which often assume well-defined degradations, tend to produce outputs with unrealistic details. Consequently, they are often less effective for applications requiring high-fidelity restoration under varying and unpredictable degradation conditions. Additionally, these methods usually require large-scale paired datasets and involve training specific models for each degradation setting, making them impractical for real-world scenarios where degradation types are diverse and dynamically changing.
\begin{figure}[t]
    \includegraphics[width=0.45\textwidth]{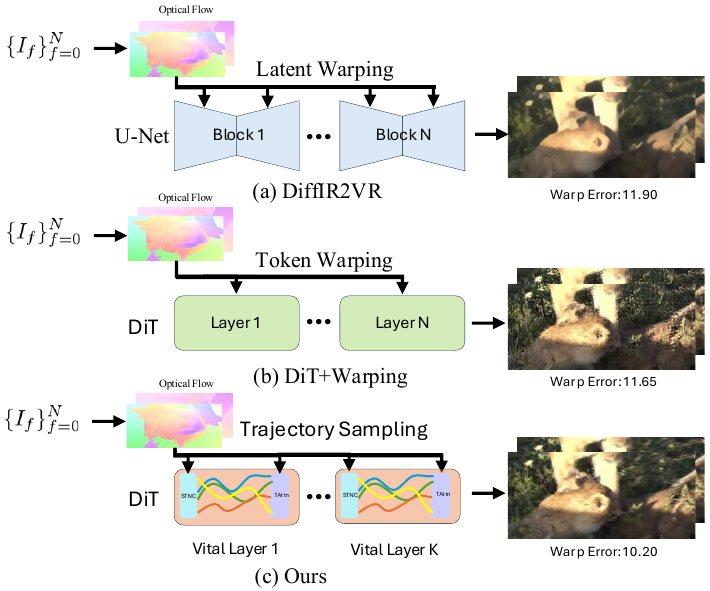}
    \caption{Model inference process comparison with previous methods. (a)  {U-Net-based zero-shot methods} \cite{yeh2024diffir2vr} rely on local convolutions, aggregating motion only within a small window and therefore exhibiting flicker and fragmented dynamics, 
    (b) {Vanilla DiT with direct flow warping} enlarges the receptive field but leaves tokens misaligned, introducing ghosting and other temporal artifacts, and (c) {DiTVR} overcomes both issues by pairing a flow-guided {spatiotemporal neighbor cache} with trajectory-aware attention} injected solely into vital layers, restoring sharp details with superior temporal consistency.
    \label{fig:motivation}
\end{figure}

Diffusion generators~\citep{ramesh2022dalle2,rombach2022high,ho2022imagen} adapt flexibly to diverse degradations and can handle multiple restoration tasks with a single network~\citep{gao2023implicit}.  Yet na\"ively extending image-based diffusers to video introduces flickering because each frame follows an independent noise path.  Recent works retrofit latent diffusion models with temporal attention or 3-D convolutions~\citep{blattmann2023align,zhou2024upscale,yang2023motion,rota2023enhancing,chen2024learning}, but these additions are computationally heavy and  rely solely on U-Net backbones that struggle with long-range dependencies.

Methods such as DiffIR2VR-Zero~\citep{yeh2024diffir2vr} and ZVRD~\citep{cao2024zero} sidestep training entirely by injecting optical-flow guidance at inference.  However, they employ U-Net architectures and use flow primarily for structural alignment, often neglecting fine-detail synthesis and global coherence.

In this paper, we propose DiTVR, a novel zero-shot video restoration framework based upon Diffusion Transformers (DiTs) that circumvents both the limited temporal receptive fields of U-Net architectures (Fig.~\ref{fig:motivation}(a)) and the spatial misalignment caused by naive optical flow warping in vanilla DiT (Fig.~\ref{fig:motivation}(b)). 
Unlike prior U-Net-based methods, which struggle to enforce temporal consistency, our approach integrates explicit optical flow guidance with targeted enhancements to vital layers that are most sensitive to temporal dynamics, maintaining structural coherence across frames while preserving fine details. This ensures high-quality video reconstructions without the need for task-specific fine-tuning.

Building on these insights, we develop DiTVR with three pivotal components.
First, a Spatiotemporal Neighbor Selection Cache mechanism reduces memory overhead by storing and retrieving key-value representations from non-overlapping spatial blocks, dynamically selected based on optical flow trajectories to enhance long-range temporal modeling. This caching strategy enables efficient adaptation to arbitrary-resolution videos and also streamlines the subsequent temporal attention process by focusing on spatially and temporally relevant information. Second, a Trajectory-Aware Attention mechanism replaces simple frame-wise attention by aligning features along motion trajectories, with particular emphasis on vital layers that are most sensitive to temporal dynamics as identified through our layer-wise analysis. This targeted approach ensures stable and temporally coherent reconstructions while mitigating ghosting artifacts. Finally, a Flow-Guided Diffusion Sampling technique extends the reverse diffusion process by integrating optical flow guidance, allowing the model to refine motion details adaptively and enforce structural consistency across frames. 

To summarize, our main contributions are threefold:
\begin{itemize}
    \item A novel zero-shot diffusion transformer framework, DiTVR, guided by optical flow trajectories for coherent and high-fidelity video restoration.
    \item We introduce a vital-layer analysis that identifies layers most sensitive to temporal dynamics and augments them with two flow-driven modules: Spatiotemporal Neighbor Cache and Trajectory-Aware Attention to improve temporal stability. Additionally, we propose a Flow-Guided Diffusion Sampler that operates during the sampling process to preserve fine-grained detail information and enforce motion consistency across frames.

    \item Extensive experiments validate the effectiveness of our method. Our model achieves state-of-the-art performance on existing zero-shot video restoration benchmarks, especially in maintaining temporal consistency.
\end{itemize}

\section{Related Work}
\label{sec:related}

\subsection{Video Restoration}
Similar to image restoration \cite{zhang2017beyond}, video restoration aims to address issues such as noise, blur, and low resolution \cite{chan2021basicvsr,chan2022basicvsr++,li2023simple,youk2024fma}. The existing video restoration methods need to be trained for every single task. However, acquiring sufficient paired training data for video restoration is significantly more difficult than for image restoration, and the reliance on predefined degradation processes \cite{liang2022recurrent,li2020mucan,liang2024vrt} limits effectiveness in real-world scenarios with unknown and diverse degradations.
Recent approaches such as Upscale-A-Video \cite{zhou2024upscale}, SATeCo \cite{chen2024learning}, and MGLD-VSR \cite{yang2023motion} extend pre-trained image diffusion models by incorporating and fine-tuning temporal layers. Unlike these trainable methods, we employ a pre-trained image diffusion model for video restoration with a training-free framework, enabling the generation of realistic details while performing the restoration.

\subsection{Diffusion Models for Image Restoration}
To improve the fidelity of details using pre-trained Stable Diffusion (SD) models, previous works \cite{wang2024exploiting,yang2023pixel,zhang2023adding,lin2023diffbir} have fine-tuned additional restoration modules. However, the latent space in these models is often susceptible to losing high-frequency image details \cite{crowson2024scalable}. Additionally, methods such as \cite{fei2023generative,wang2023zero,kawar2022denoising} have adopted more efficient approaches by constraining the sampling process of pre-trained diffusion models. Nevertheless, directly applying these techniques to video restoration often results in temporal inconsistencies, specifically evident in the preservation of local details. Unlike other approaches, the proposed DiTVR is compatible with pre-trained diffusion models with DiTs  \cite{peebles2023scalable} in pixel space and incorporates a temporal guided constraint in the reverse diffusion process, ensuring better temporal consistency and more accurate restoration of video details.

\subsection{Diffusion Models for Zero-shot Video Tasks}
Along with the development of powerful pre-trained generative diffusion models, existing video methods have shifted towards zero-shot frameworks, which leverage the strong generative priors of these models and primarily aim to address the flickering artifacts caused by temporal inconsistencies.
\cite{geyer2023tokenflow,yang2023rerender} propose a two-stage zero-shot video editing framework that involves keyframe editing followed by full video propagation to maintain temporal consistency across frames.
Alternatively, various methods such as token merging \cite{li2024vidtome}, random shuffling \cite{feng2024wave}, and latent warping \cite{bao2023latentwarp} have demonstrated the ability to produce remarkable editing outcomes without the need for additional training. Building on these techniques, DiffIR2VR-Zero \cite{yeh2024diffir2vr} incorporates hierarchical latent warping and refines token merging by leveraging flow correspondence and spatial information, minimizing blurring artifacts in challenging video restoration tasks. 

\section{Preliminaries}
\label{sec:preliminaries}

\begin{figure*}[t!]
    \centering
    \includegraphics[width=0.6\textwidth]{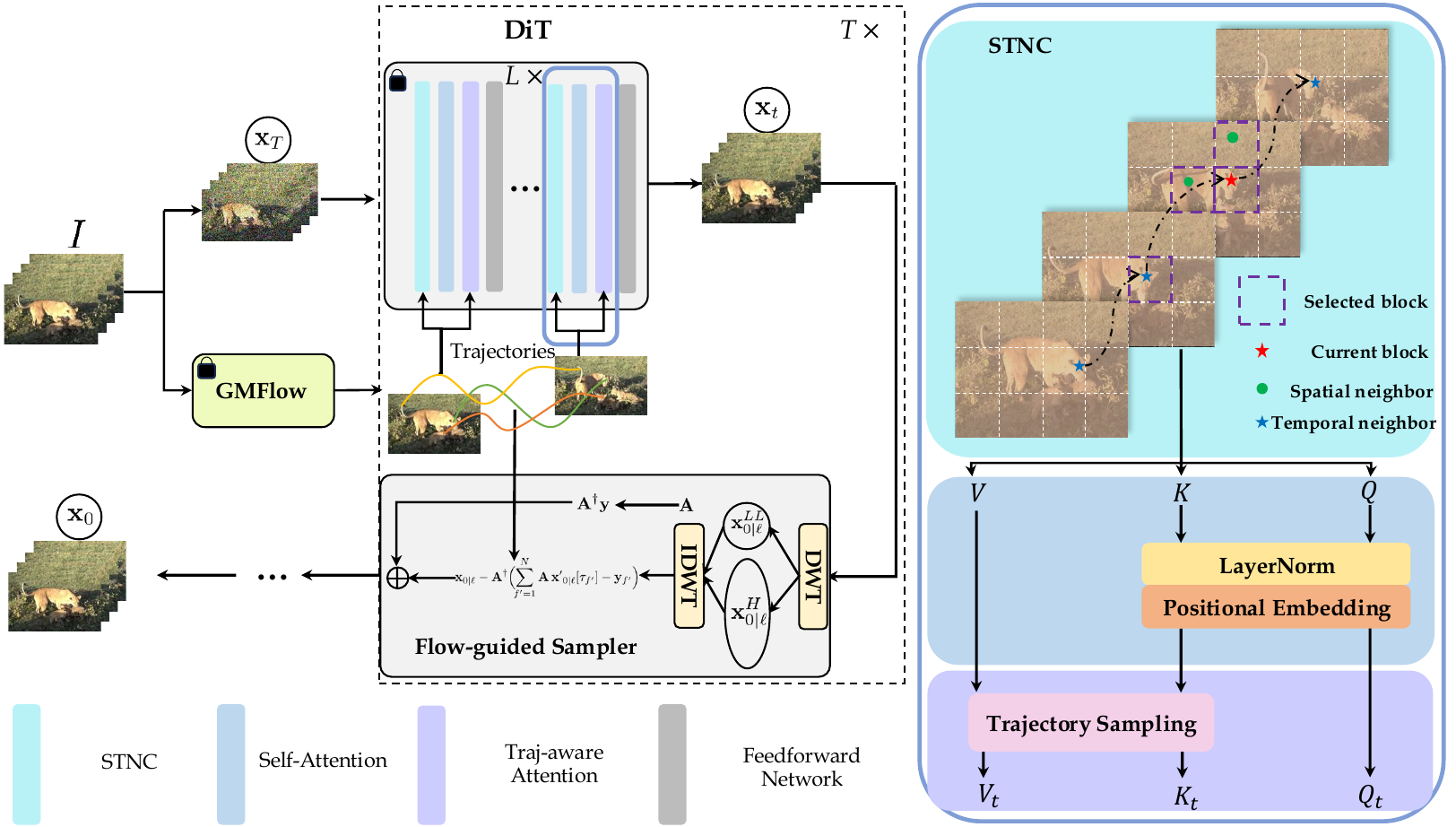}
    \caption{{\bf Overall pipeline of the proposed zero-shot video restoration framework}. 
    }
    \label{fig:pipeline}
\end{figure*}
\subsection{Diffusion Models}
\label{pre:diffusion}
We firstly briefly review the fundamental concepts behind diffusion models (DDPMs)~\cite{sohl2015deep,ho2020denoising}. Gaussian diffusion models assume a forward noising process that gradually adds noise to the real data $x_0$: 
\begin{equation}
q(x_t|x_0) = \mathcal{N}(x_t; \sqrt{\bar{\alpha}_t}x_0, (1 - \bar{\alpha}_t)\mathbf{I}),
\end{equation}
where constants $\bar{\alpha}_t$ are the predefined hyperparameter for the timestep $t$. A clean image $x_0$ can be diffused by Gaussian noises in T steps $x_t = \sqrt{\bar{\alpha}_t}x_0 + \sqrt{1 - \bar{\alpha}_t} \epsilon_t$, where $\epsilon_t \sim \mathcal{N}(0,\mathbf{I})$. Diffusion models are trained to learn the reverse process, which aims to invert the forward process corruptions: 
\begin{equation}
p_\theta(x_{t-1}|x_t) = \mathcal{N}(\mu_\theta(x_t), \Sigma_\theta(x_t)),
\end{equation}
where $\mu_\theta(x_t)$ and $\Sigma_\theta(x_t)$ are computed by the denoising model $\epsilon_\theta$. Note that our model $\epsilon_\theta$ is similar to DiT~\cite{peebles2023scalable}, which differs from other zero-shot video restoration methods based on U-Net~\citep{ronneberger2015unet} due to its superior efficacy and scalability.
\subsection{Optical Flow Trajectory Sampling}
\label{pre:traj}

{To establish reliable trajectories across video frames, we employ bi-directional optical flow estimation using the GMFlow model~\citep{xu2022gmflow}, which captures both forward $\mathbf{f}^{\text{fwd}}_{f}$ and backward $\mathbf{f}^{\text{bwd}}_{f}$ flow fields between consecutive frames 
.} 
The forward flow maps each pixel $(u_f, v_f)$ in frame $f$ to its corresponding position in frame $f+1$ as:
\begin{equation} 
{ 
(u_{f+1}, v_{f+1}) = (u_f, v_f) + \mathbf{f}^{\text{fwd}}{f}(u_f, v_f), } \end{equation}
{where $\mathbf{f}^{\text{fwd}}{f}(u_f, v_f)$ denotes the forward optical flow displacement at $(u_f, v_f)$. The backward flow $\mathbf{f}^{\text{bwd}}_{f}$ provides a complementary mapping from frame $f+1$ back to frame $f$, thus enabling bidirectional trajectory validation.}

Unlike U-Net architectures that perform repeated downsampling, our Transformer-based framework maintains direct spatial representation. We downsample optical flow fields only once to align with RGB pixel resolution, preventing information loss while enabling precise trajectory tracking.


Given the downsampled optical flow, we initialize block positions in the first frame and propagate them across subsequent frames to construct a trajectory set $\{\tau_k\}_{k=1}^K$. To ensure temporal consistency, each trajectory undergoes bidirectional validation. Specifically, a candidate trajectory extending from $(u_f, v_f)$ in frame $f$ to $(u_{f+1}, v_{f+1})$ in frame $f+1$ is accepted only if the backward-mapped position $(u_f^{\text{bwd}}, v_f^{\text{bwd}})$ remains spatially consistent with the original:
\begin{equation}
\tau_k = \begin{cases} (f{+}1, u_{f+1}, v_{f+1}), & \text{if } (u_f^{\text{bwd}}, v_f^{\text{bwd}}) \in \mathcal{N}_1(u_f, v_f) \\ (-1, -1, -1), & \text{otherwise} \end{cases}
\end{equation}
Here, $(u_f^{\text{bwd}}, v_f^{\text{bwd}})$ is obtained by warping $(u_{f+1}, v_{f+1})$ back to frame $f$ using $\mathbf{f}^{\text{bwd}}_{f}$ and $\mathcal{N}_1(u_f, v_f)$ denotes the 1-pixel neighborhood of the spatial coordinate $(u_f, v_f)$. If the backward-projected position deviates beyond a small threshold, the trajectory is discarded to prevent drift and misalignment. This validation step ensures that only stable, non-occluded motion paths contribute to our temporal modeling. The resulting trajectories serve as a foundation for motion-aware attention mechanisms, enabling robust temporal coherence in subsequent processing stages.

\section{Methodology}
\label{sec:method}

Given a degraded video of length $N$, denoted by $\{I_f\}_{f=1}^{N}$, our goal is to restore it into a high-quality video $\{I'_f\}_{f=1}^{N}$ using a pre-trained image diffusion model. Unlike prior methods~\citep{yeh2024diffir2vr, cao2024zero} that rely on U-Net architectures, our approach employs a Diffusion Transformer (DiT) pre-trained in pixel space. We leverage optical flow trajectories from the input video $\{I_f\}_{f=1}^{N}$ (Sec.~\ref{sec:cache} to Sec.~\ref{sec:sampler}) as the basis for our three main components (Fig.~\ref{fig:pipeline}): (1) a spatiotemporal neighbor selection cache (STNC) mechanism (Sec.~\ref{sec:cache}) for efficient token management, (2) trajectory-aware attention (Sec.~\ref{sec:attn}) to enforce temporal consistency, and (3) a flow-guided diffusion sampler (Sec.~\ref{sec:sampler}) to preserve temporal coherence across frames.

\subsection{STNC Mechanism}
\label{sec:cache}

While previous methods, such as Unidirectional Block Attention~\citep{yang2024inf}, select neighboring blocks based solely on spatial adjacency, they overlook inter-frame motion patterns. To address this limitation, we introduce a Spatiotemporal Neighbor Selection Cache (STNC) mechanism that integrates optical flow trajectories for dynamic token selection. Unlike conventional approaches that rigidly select adjacent blocks, STNC identifies truly relevant neighbors based on motion correspondences across frames, ensuring both spatial and temporal coherence for superior video restoration.

Let each frame \(I_f\) be first encoded into hidden states \(\mathbf{h}_f \in \mathbb{R}^{H \times W \times D}\), which are then partitioned into non-overlapping blocks \(\{\mathbf{h}_f^{(p,q)}\}\), with \((p,q)\) indexing each block. Within each layer, every block is influenced by its spatial neighbors, specifically the blocks to its left and above, denoted as \(\mathbf{h}_{f,\text{s}}^{(p,q)}\) (shown via green color in the upper-right corner of Fig.~\ref{fig:pipeline}), ensuring {intra-frame} consistency. To extend this spatial reasoning across frames and capture {inter-frame} motion, we define \(\tau_f\) as the optical flow field that maps pixels from frame \(f\) to frame \(f-1\). For a given block \(\mathbf{h}_f^{(p,q)}\), we compute the number of its pixels that map to each block in the previous frame, \(\mathbf{h}_{f-1}^{(p',q')}\), via \(\tau_f\). Let \(n_{p',q'}\) represent the count of pixels in \(\mathbf{h}_f^{(p,q)}\) that flow into \(\mathbf{h}_{f-1}^{(p',q')}\). The {inter-frame} temporal neighbor for the block \(\mathbf{h}_f^{(p,q)}\) is then selected by choosing the block \(\mathbf{h}_{f-1}^{(p',q')}\) that receives the maximum number of pixel mappings, as shown via the temporal neighbor in the purple box and formalized as:
\begin{equation}
    \mathbf{h}_{f,\text{t}}^{(p', q')} = \underset{(p',q')}{\arg\max} \; n_{p',q'},
\end{equation}
This approach ensures that each block \(\mathbf{h}_f^{(p,q)}\) is associated with the most relevant {temporal} neighbor \(\mathbf{h}_{f-1}^{(p',q')}\), preserving both {intra-frame} and {inter-frame} coherence. By caching only a small, relevant set of spatial and temporal neighbors rather than entire frames, we significantly reduce memory usage while maintaining the necessary spatiotemporal context for effective video restoration.


\begin{figure}[h]
    \centering
    \includegraphics[width=0.45\textwidth]{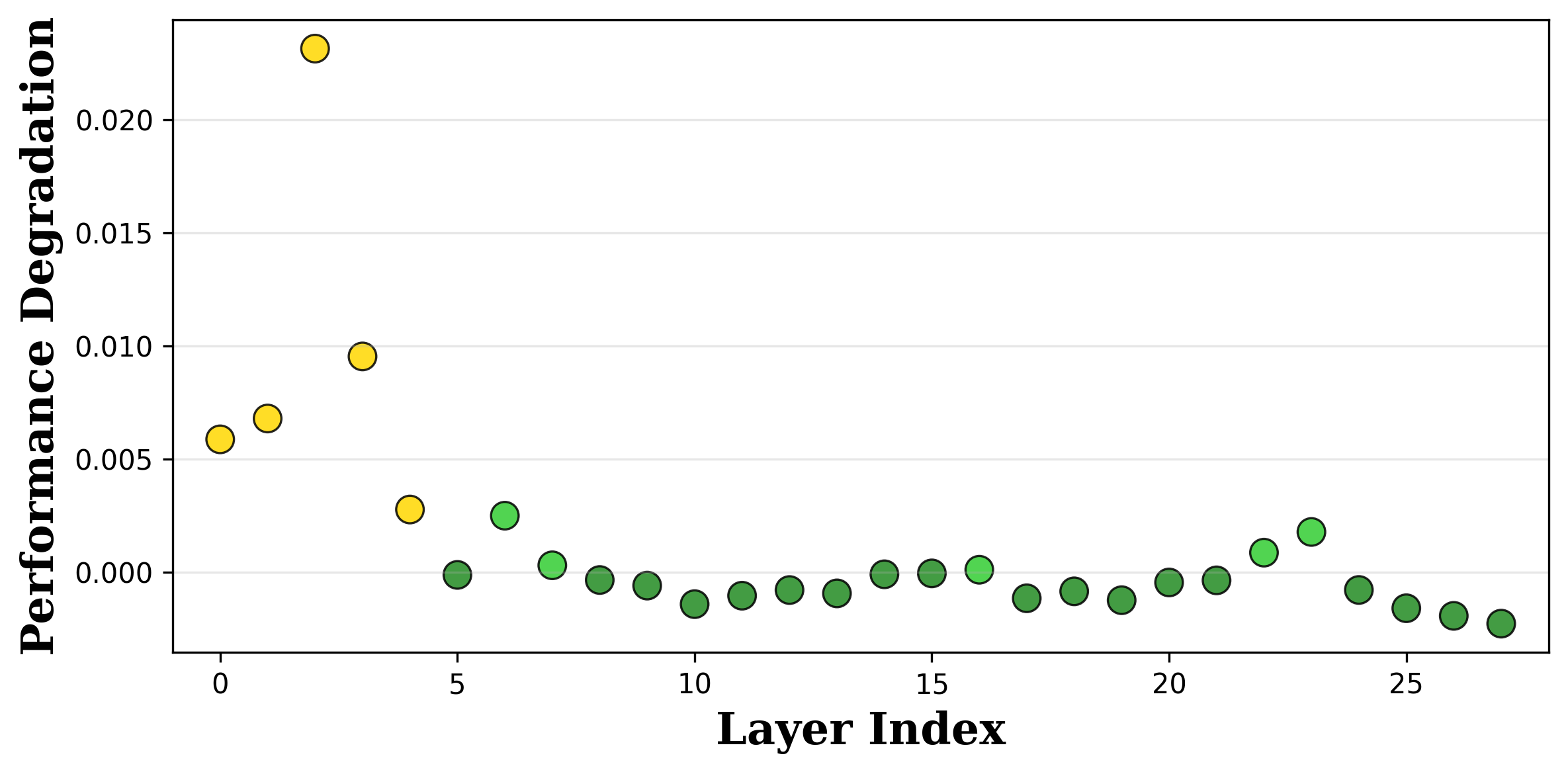}
    \caption{\textbf{Layer-wise analysis of temporal performance degradation.} We measure the effect of removing each layer by calculating the performance degradation in temporal consistency metrics. Higher values indicate that removing the layer significantly impacts temporal coherence, while lower values suggest minimal influence on video restoration quality. Yellow points highlight the most influential layers (vital layers) that are critical for maintaining temporal consistency. As shown, these vital layers are distributed across the transformer architecture rather than concentrated in specific regions, demonstrating the importance of preserving temporal-sensitive components throughout the network depth.}
    \label{fig:vital}
\end{figure}
\subsection{Trajectory-aware Attention}
\label{sec:attn}

Building on the adaptive neighbor selection, we introduce two-stage {Trajectory-aware Attention,} which integrates self-attention within the blocks and cross-attention among the selected spatial/temporal neighbors to enforce inter-frame consistency. As demonstrated in our layer-wise analysis (Fig.~\ref{fig:vital}), certain transformer layers exhibit significantly higher sensitivity to temporal dynamics, with these vital layers distributed throughout the network architecture rather than concentrated in specific regions. Our trajectory-aware attention specifically targets these temporal-critical layers to maximize the preservation of inter-frame coherence while maintaining computational efficiency.
\begin{table*}[h]
\centering
\caption{{\bf Quantitative evaluation}. Comparison of the video super-resolution results on DAVIS~\citep{davis}, SPMC~\citep{tao2017detail} and Vid4~\citep{nah2019ntire} datasets. The best and second performances are marked in red and blue, respectively. \textit{The dashed line separates supervised methods (above) from zero-shot methods (below).}}
\resizebox{\textwidth}{!}{
\begin{tabular}{c|ccccc}
\hline
Methods         & PSNR$\uparrow$ & SSIM$\uparrow$ & LPIPS$\downarrow$ & WE$\left(\times 10^{-3}\right)\downarrow$ & FSim$\uparrow$\\ \hline
FMA-Net~\cite{youk2024fma}         &  {\color{red}33.51}/{\color{red}31.18}/{\color{red}31.16}    &  {\color{red}0.8231}/{\color{red}0.6182}/{\color{red}0.7726}    &   {\color{blue}0.1580}/{\color{blue}0.1843}/{\color{blue}0.2347}    &  14.30/2.208/{\color{red}3.839}  & 0.9583/{\color{blue}0.9925}/{\color{blue}0.9837}  \\ 
\cdashline{1-6}
Inf-DiT~\citep{yang2024inf}(per frame) &  29.12/29.18/28.37    &  0.4356/0.3465/0.1865    &   0.3967/0.2977/0.4836    &  15.64/6.040/10.99  & 0.9566/0.9850/0.9704  \\
Inf-DiT~\citep{yang2024inf}(warping) & 31.27/30.03/29.15 & 0.4682/0.3567/0.1916 & 0.3693/0.2892/0.4705 & 13.82/4.956/9.124 & 0.9615/0.9889/0.9751 \\
DiffIR2VR-Zero~\citep{yeh2024diffir2vr}  &  32.28/30.56/29.78    &  0.7306/0.5671/0.5239    & 0.1623/0.2240/0.2825      & {\color{blue}10.95}/2.357/5.426  & {\color{blue}0.9648}/0.9922/{\color{red}0.9856}  \\
Ours            &   {\color{blue}33.29}/{\color{blue}30.94}/{\color{blue}30.36}   &  {\color{blue}0.7870}/{\color{blue}0.6025}/{\color{blue}0.6401}    &   {\color{red}0.1216}/{\color{red}0.1570}/{\color{red}0.2236}    &  {\color{red}10.90}/2.144/{\color{blue}5.338}  & {\color{red}0.9699}/{\color{red}0.9945}/{\color{red}0.9856}  \\ \hline
\end{tabular}
}

\label{tab:sr}
\end{table*}

\begin{figure*}[h]
    \centering
    \includegraphics[width=0.7\textwidth]{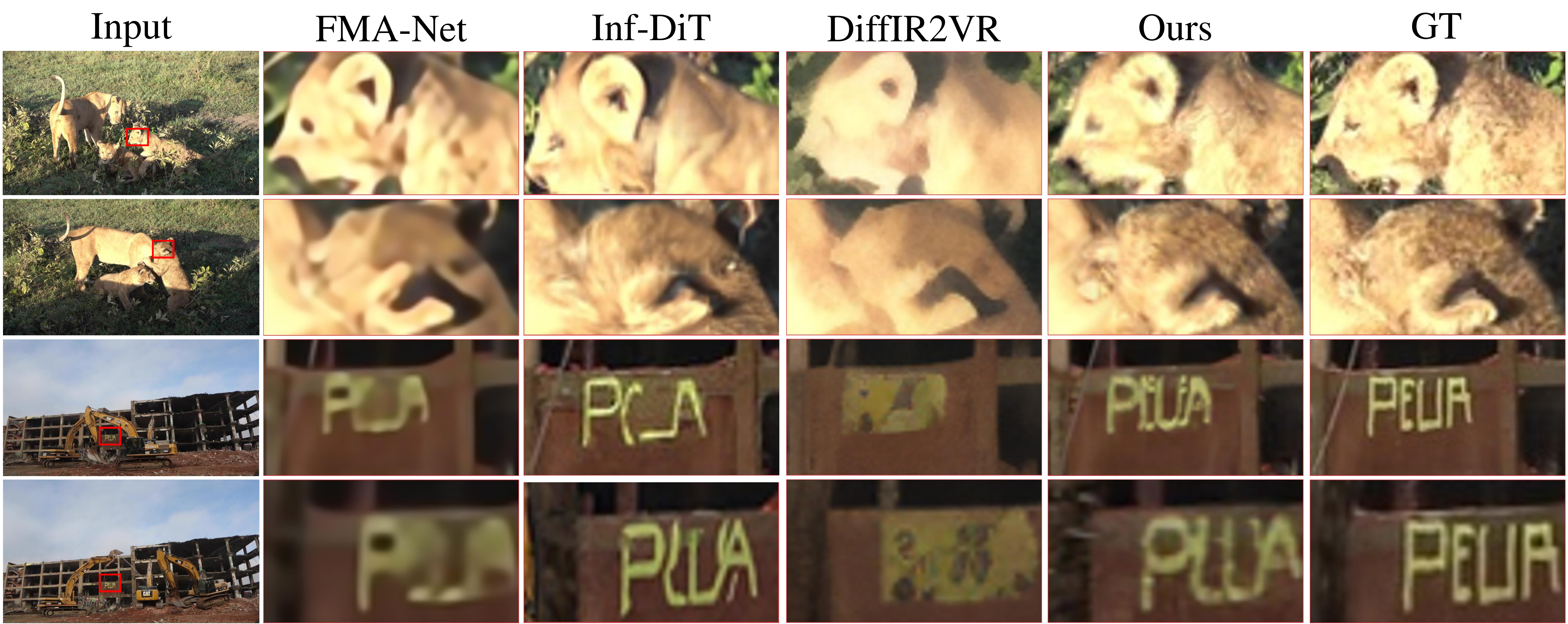}
    \caption{{\bf Qualitative comparisons on 4$\times$ video super-resolution}. While applying models like Inf-DiT to individual frames can generate realistic details, they often lack temporal consistency. In contrast, our method achieves both realistic and consistent results across frames without requiring additional training. }
    \label{fig:sr}
\end{figure*}
{Let $\mathbf{h}_f^{(p,q),l}$ denote the hidden state for the block at layer $l$.} To construct the query vector, we project the current block representation as $\bm{Q} = \mathbf{h}_f^{(p,q),l}\bm{W}^Q$. For the key $\bm{K}$ and value $\bm{V}$ vectors, we concatenate the hidden states from three sources: the current block, its spatial neighbors, and its most relevant temporal neighbor as determined by optical flow: 
\begin{align}
\label{eq:key}
\bm{K} &= \left[\mathbf{h}_f^{(p,q),l} + \bm{P}_0;\ \mathbf{h}_{f,\text{s}}^{l} + \bm{P}_{\text{s}};\ \mathbf{h}_{f,\text{t}}^{l} + \bm{P}_{\text{t}} \right] \bm{W}^K \\
\label{eq:value}
\bm{V} &= \left[\mathbf{h}_f^{(p,q),l};\ \mathbf{h}_{f,\text{s}}^{l};\ \mathbf{h}_{f,\text{t}}^{l} \right] \bm{W}^V,
\end{align}
where $[\cdot]$ denotes concatenation, \(\bm{P}\) are block-level positional encodings, and subscripts \(\text{s},\text{t}\) signify “spatial” and “temporal” neighbors, respectively. We then apply scaled dot-product attention:
\begin{equation}
\label{eq:attention}
\bm{c^{l}} = \text{Softmax}\!\Big(\!\frac{\bm{Q}\,\bm{K}^{\top}}{\sqrt{d}}\Big)\,\bm{V},
\end{equation}
where \(d\) is the feature dimensionality. Thus, the first stage ensures that the hidden representation is enriched with spatially relevant information while maintaining efficiency by limiting the number of queried blocks.

To further enhance temporal consistency, we introduce an additional cross-attention mechanism that directly aggregates information from prior frames along flow-based trajectories. For each flow trajectory $\tau_f$, we gather corresponding blocks from previous frames $f-1, f-2,$ \textit{etc,}.\ constructing trajectory specific matrices $\bm{K}_{\text{t}}, \bm{V}_{\text{t}}$ leading to:

\begin{equation}
\label{eq:temporal_attention}
\bm{c}^{l}_{\text{t}} = \text{Softmax}\!\Big(\!\frac{\bm{Q}_{\text{t}}\big(\bm{K}_{\text{t}}[\tau_f]\big)^\top}{\sqrt{d}}\Big)\,\bm{V}_{\text{t}}[\tau_f].
\end{equation}
Finally, the updated hidden state is computed as \(\mathbf{h}_f^{(p,q),l+1} = \text{FFN}(\bm{c}^{l}_{\text{t}})\), ensuring that each block benefits from both spatially and temporally relevant information. By aligning features along flow trajectories, trajectory-aware attention mechanism avoids ghosting artifacts and improves motion coherence, leading to superior zero-shot video restoration.
\begin{table*}[]
\centering
\caption{{\bf Quantitative evaluation}. Comparison of the blind video super-resolution results on DAVIS~\citep{davis}, SPMC~\citep{tao2017detail}, and Vid4~\citep{nah2019ntire} datasets. The results are presented as ``DAVIS/SPMC/Vid4'', with a single ``/'' separating the datasets.} 
\resizebox{\textwidth}{!}{
\begin{tabular}{c|ccccc}
\hline
Methods         & PSNR$\uparrow$ & SSIM$\uparrow$ & LPIPS$\downarrow$ & WE$\left(\times 10^{-3}\right)\downarrow$ & FSim$\uparrow$ \\ \hline
FMA-Net~\cite{youk2024fma} & 29.97/29.03/{\color{red}31.00}       & 0.6448/0.4670/{\color{red}0.7559}     &  0.4930/0.6238/{\color{blue}0.2347}    &  15.61/13.88/{\color{red}3.839} & 0.9608/0.9698/{\color{blue}0.9863}       \\
\cdashline{1-6}
Inf-DiT~\citep{yang2024inf}(per frame) &   28.89/28.53/28.30   &   0.4259/0.3135/0.1734  &   0.4628/0.5017/0.5141    & 17.58/19.81/22.57  &  0.9566/0.9576/0.9471  \\
Inf-DiT~\citep{yang2024inf}(warping) & 29.45/29.12/28.95 & 0.4580/0.3420/0.1890 & 0.4320/0.4650/0.4785 & 13.23/14.84/10.92 & 0.9612/0.9635/0.9548 \\
DiffIR2VR-Zero~\citep{yeh2024diffir2vr}  & {\color{blue}30.68}/{\color{red}29.39}/29.77     &  {\color{blue}0.6583}/{\color{blue}0.4946}/0.5301    & {\color{red}0.2857}/{\color{blue}0.4526}/0.2977     & {\color{blue}11.90}/{\color{blue}9.072}/4.250  & {\color{blue}0.9623}/{\color{blue}0.9780}/{\color{red}0.9893}    \\
Ours            &  {\color{red}30.69}/{\color{red}29.39}/{\color{blue}30.41}    &   {\color{red}0.6710}/{\color{red}0.4950}/{\color{blue}0.6326}  &    {\color{blue}0.3623}/{\color{red}0.4449}/{\color{red}0.2115}   &  {\color{red}10.20}/{\color{red}9.022}/{\color{blue}4.004} & {\color{red}0.9697}/{\color{red}0.9820}/0.9688 \\ \hline
\end{tabular}
}

    \label{tab:bsr}

\end{table*}
\subsection{Flow-guided Diffusion Sampler}
\label{sec:sampler}

Although the cache and attention modules keep features coherent inside the DiT, a vanilla DDIM/DDNM sampler still propagates noise frame-by-frame, resulting in subtle flicker.  
We therefore augment the sampling process with a {wavelet-domain, flow-aware correction} that acts only on the information most critical for temporal stability.

{\textbf{Step 1: Wavelet split.}}\;
For every diffusion step $\ell$ (distinct from the frame index $f$), we decompose the current DiT prediction $\mathbf{x}_{0|\ell}$ into low- and high-frequency sub-bands via a discrete wavelet transform (DWT):
\begin{equation}
\bigl\{\mathbf{x}_{0|\ell}^{ij}\bigr\}_{i,j \in\{L,H\}} 
       \;=\; \operatorname{DWT}\!\bigl(\mathbf{x}_{0|\ell}\bigr),
\end{equation}
where the $L$, low-frequency band captures structure and motion, while the $H$, high-frequency band contains fine details.

{\textbf{Step 2: Low-Frequency data fidelity.}}\;
Following the range/null-space philosophy of DDNM \cite{wang2022zero}, we apply the degradation operator $\mathbf{A}$ {only} to the low-frequency part.  
Let $\mathbf{A}^\dagger$ be its pseudo-inverse and $\mathbf{y}_f$ the observed LQ frame; the corrected estimate is
\begin{equation}
\hat{\mathbf{x}}_{0|\ell} \;=\;
      \mathbf{x}_{0|\ell} \;-\;
      \mathbf{A}^\dagger \bigl(\mathbf{A}\,\mathbf{x'}_{0|\ell} - \mathbf{y}_f \bigr),
      \label{eq:ddnm_plus_video}
\end{equation}
which enforces consistency in the range space while leaving high-band detail to the DiT prior.

\vskip2pt\noindent\textbf{Step 3: Flow-guided residual alignment.}\;
To suppress inter-frame drift, we further average residuals along the forward optical-flow trajectories $\tau_f$ across a temporal window of $N$ neighbours:
\begin{equation}
\hat{\mathbf{x}}_{0|\ell} \;=\;
\mathbf{x}_{0|\ell} \;-\;
\mathbf{A}^\dagger
\Bigl(
\sum_{f'=1}^{N}
      \mathbf{A}\,\mathbf{x'}_{0|\ell}\bigl[\tau_{f'}\bigr]
      - \mathbf{y}_{f'}
\Bigr).
\label{eq:flow_guided_correction}
\end{equation}
Blending range-space corrections along reliable motion paths removes flicker without blurring textures, so every diffusion step yields a temporally smoother video.

\section{Experiments}
\label{sec:experiments}


\paragraph{Implementation Details.}
Our experiments are conducted on an NVIDIA RTX 4090 GPU. We apply our method to the image-based Diffusion Transformer model Inf-DiT~\citep{yang2024inf}, though it can be readily implemented with other Diffusion Transformer models as well.
\paragraph{Testing Datasets.} For video super-resolution, we evaluate on Vid4~\citep{nah2019ntire}, SPMCS~\citep{tao2017detail} and DAVIS~\citep{davis} testing sets, with 4$\times$ downsample scales. For blind video super-resolution, we 
follow the degradation pipeline of RealBasicVSR~\cite{chan2021basicvsr} to achieve degraded 4$\times$ videos. For video denoising, we evaluate on DAVIS and SPMCS~\cite{nah2019ntire} with different noise levels (std. $=$ 50, 75, and is uniformly sampled from the range [50, 100]). 
\paragraph{Evaluation Metrics.}
We select five measurements to evaluate the restoration and enhancement quality. Besides the common video quality metrics PSNR, SSIM, and LPIPS, we utilize Warping Error (WE)~\citep{lai2018learning,ceylan2023pix2video} and Frame Similarity (Fsim) 
metrics to evaluate temporal consistency.
\subsection{Comparisons with State-of-the-art Methods}
To verify the performance of our method, we compare DiTVR with trainable state-of-the-art video restoration methods including FMA-Net~\citep{youk2024fma} for video super-resolution and Shift-Net~\citep{li2023simple} for video denoising. We also compare our method with the per-frame restoration method Inf-DiT~\citep{yang2024inf} and a zero-shot restoration method DiffIR2VR-Zero~\citep{yeh2024diffir2vr}. To establish a more comprehensive baseline, we also implement Inf-DiT with optical flow warping (Inf-DiT(warping)) by adapting the same warping mechanism used in DiffIR2VR, which provides temporal consistency without requiring additional training.

\begin{figure}[t]
    \centering
\includegraphics[width=0.45\textwidth]{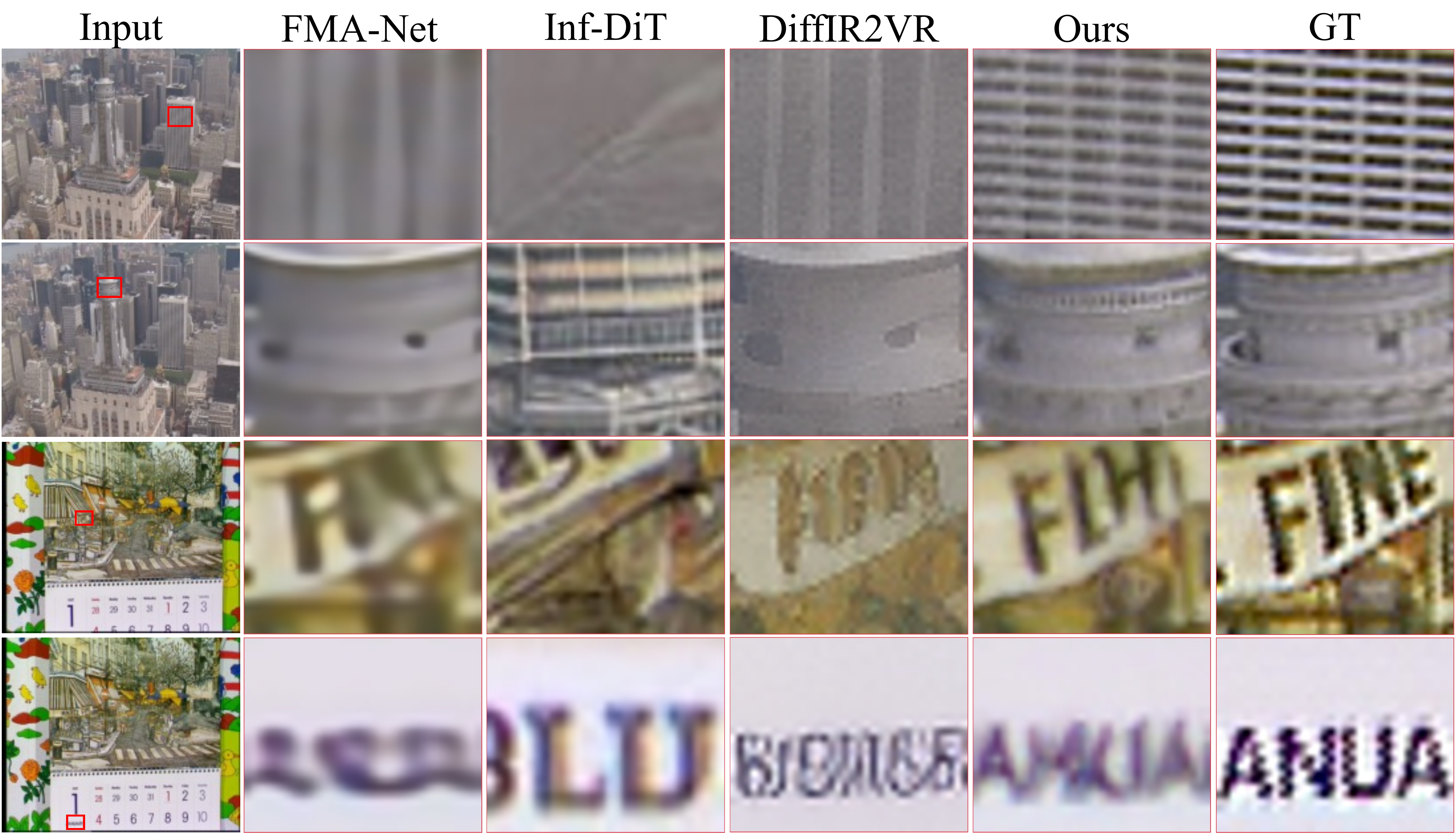}
    \caption{{\bf Qualitative comparisons on 4$\times$ blind video super-resolution on Vid4 dataset}. In the magnified patches, our method delivers sharper and more consistent reconstruction.}
    \label{fig:bsr}
\end{figure}
\begin{figure*}[h]
    \centering
    \includegraphics[width=0.65\textwidth]{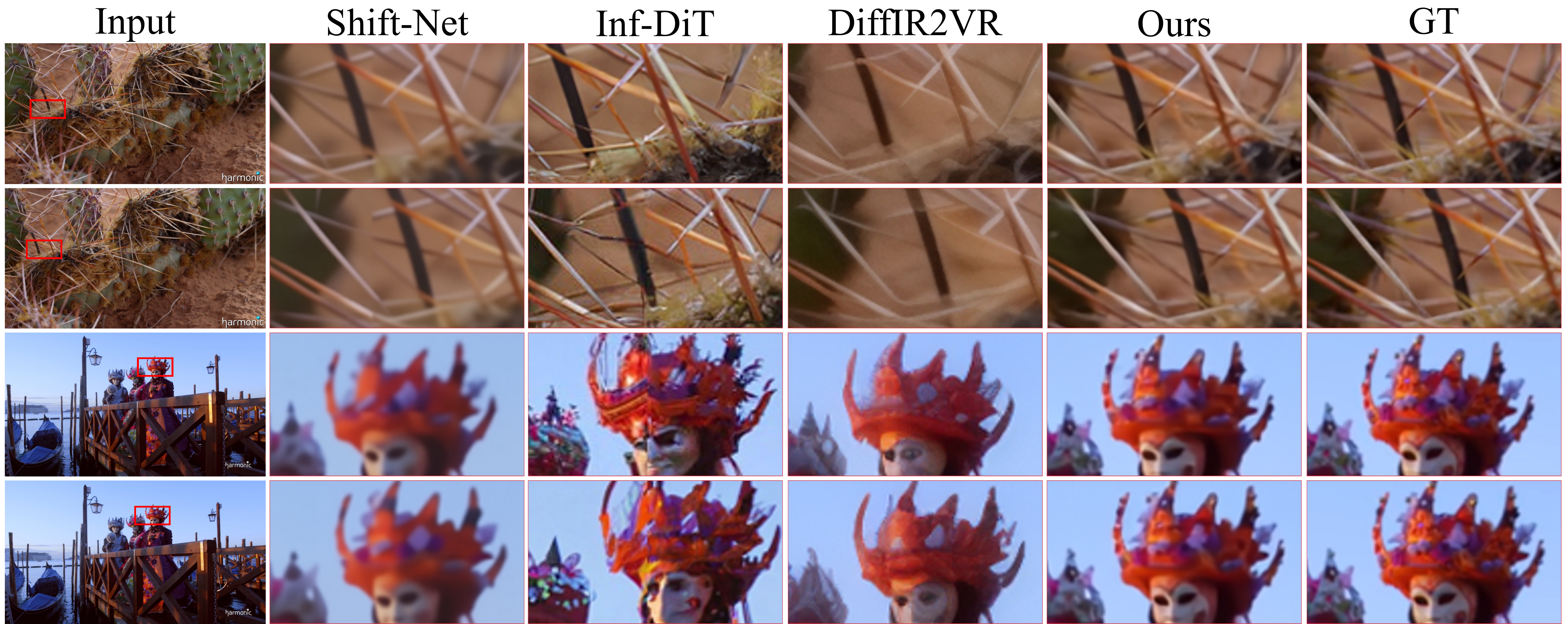}
    \caption{{\bf Qualitative comparison on video denoising ($\sigma=75$)}. Our method effectively denoises and generates detailed frames while preserving temporal consistency.}
    \label{fig:dn}
\end{figure*}
\begin{table*}[h]
\centering
\caption{{\bf Quantitative Evaluation}. Comparison of video denoising results on the DAVIS~\citep{davis} and SPMC~\citep{tao2017detail} datasets with noise levels of $\sigma = 50$ and $75$. The results are presented as ``DAVIS/SPMC'', using a single ``/'' to split different datasets.} 
\resizebox{0.85\textwidth}{!}{
\begin{tabular}{c|c|ccccc}
\hline
Methods         & $\sigma$ & PSNR$\uparrow$ & SSIM$\uparrow$ & LPIPS$\downarrow$ & WE$\left(\times 10^{-3}\right)\downarrow$ & FSim$\uparrow$ \\ \hline
Shift-Net~\citep{li2023simple} & 50 & {\color{red}33.99}/{\color{red}36.18} & {\color{red}0.8502}/{\color{red}0.9538} & {\color{red}0.0824}/{\color{red}0.0356} & 16.98/2.632 & 0.9638/0.9930\\
 & 75 & {\color{red}33.48}/{\color{red}34.77} & {\color{red}0.8355}/{\color{red}0.9333} & {\color{red}0.1130}/{\color{red}0.0588} & 16.81/2.627 &  0.9644/0.9931 \\
\cdashline{1-7}
Inf-DiT~\citep{yang2024inf} (per frame) & 50 & 28.89/29.13 & 0.4261/0.3340 & 0.4627/0.3220 & 17.62/10.26 & 0.9544/0.9755\\
 & 75 & 28.51/29.01 & 0.4167/0.3344 & 0.4654/0.3366 & 18.54/11.22 & 0.9501/0.9739 \\
Inf-DiT~\citep{yang2024inf} (warping) & 50 & 30.15/30.42 & 0.4582/0.3668 & 0.4315/0.2985 & 15.84/8.756 & 0.9587/0.9798\\
 & 75 & 29.78/30.28 & 0.4485/0.3671 & 0.4341/0.3121 & 16.72/9.583 & 0.9543/0.9782 \\
DiffIR2VR-Zero~\citep{yeh2024diffir2vr}  & 50 & 31.98/30.69 & 0.7582/0.6820 & 0.1596/0.2074 & {\color{blue}10.45}/{\color{blue}2.237} & {\color{blue}0.9710}/{\color{red}0.9951} \\
 & 75 & 31.54/30.57 & 0.7406/0.6776 & 0.1924/0.2269 & {\color{blue}10.71}/{\color{blue}2.487} & {\color{blue}0.9705}/{\color{blue}0.9946} \\
Ours            & 50 & {\color{blue}32.18}/{\color{blue}31.70} & {\color{blue}0.8197}/{\color{blue}0.8542} & {\color{blue}0.1495}/{\color{blue}0.1409} & {\color{red}9.243}/{\color{red}1.767} & {\color{red}0.9738}/{\color{blue}0.9944}  \\
 & 75 & {\color{blue}31.96}/{\color{blue}30.61} & {\color{blue}0.7935}/{\color{blue}0.7865} & {\color{blue}0.1870}/{\color{blue}0.2058} & {\color{red}8.717}/{\color{red}1.614} & {\color{red}0.9768}/{\color{red}0.9957} \\ \hline
\end{tabular}
}

\label{tab:dn}

\end{table*}

\subsubsection{Video Super-Resolution}
As shown in Table~\ref{tab:sr}, the supervised method FMA-Net~\citep{youk2024fma} achieves the highest PSNR and SSIM scores but performs poorly on warping error (DAVIS and SPMCS), indicating challenges in handling complex video motion. DiffIR2VR-Zero~\citep{yeh2024diffir2vr}, while capable of producing temporally consistent results, tends to generate overly smooth and blurry outputs, compromising perceptual quality.
In contrast, our method enhances temporal consistency while preserving the fine details inherent to the original diffusion model. Fig.~\ref{fig:sr} presents a visual comparison of two cases from the DAVIS dataset. Due to its regression-based training, FMA-Net struggles to generate realistic details. Inf-DiT, despite its ability to capture intricate local textures, suffers from a lack of temporal coherence, resulting in inconsistencies across frames.  
DiffIR2VR-Zero maintains global consistency throughout the video but fails to resolve detail loss, leading to perceptually flat and flickering outputs. In comparison, our zero-shot video restoration framework transforms low-quality inputs into high-resolution outputs by effectively balancing temporal alignment and spatial detail preservation, producing high-quality, temporally consistent video frames.
\paragraph{ Blind Video Super-Resolution.}
We also evaluate the proposed model on the blind super-resolution task to demonstrate its efficacy.
As shown in Table~\ref{tab:bsr}, FMA-Net achieves the highest PSNR and SSIM on Vid4 dataset but suffers from poor perceptual quality, as indicated by its higher LPIPS score and inconsistent outputs shown in Fig.~\ref{fig:bsr}.
Inf-DiT produces visually appealing outputs but frequently introduces hallucinated details, compromising fidelity.
 DiffIR2VR-Zero \citep{yeh2024diffir2vr} generates sharp textures but fails to maintain temporal consistency, as reflected by its higher warping error (WE) across all the datasets. In contrast, our method not only enhances temporal consistency  (lowest WE and highest FSim) on DAVIS and SPMC but also achieves the lowest LPIPS scores on the SPMC and Vid4 datasets, delivering visually sharp and coherent outputs that demonstrate a superior balance between perceptual quality and temporal coherence.

\begin{table}[]
\centering
\caption{\textbf{Ablation study} evaluating the impact of the trajectory-aware attention (TAttn), spatiotemporal neighbor cache (STNC), and flow-guided sampler (FS) on video restoration. A checkmark (\checkmark) indicates the component is included, and a dash (--) indicates it is omitted.}
\resizebox{0.45\textwidth}{!}{

\begin{tabular}{cccc|cccc}
\toprule
\multicolumn{4}{c|}{Components} & \multicolumn{4}{c}{Metrics} \\
\cmidrule{1-4} \cmidrule{5-8}
Warping & TAttn & STNC & FS & PSNR $\uparrow$ & SSIM $\uparrow$ & LPIPS $\downarrow$ & WE $\downarrow$ \\
\midrule
-- & -- & -- & -- & 28.37 & 0.1865 & 0.4836 & 10.99 \\
\checkmark & -- & -- & -- & 29.15 & 0.1916 & 0.4705 & 9.124 \\
-- & \checkmark & -- & -- & 30.18 & 0.6187 & 0.2187 & 6.739 \\ 
-- & \checkmark & 1 & -- & 30.31 & 0.6345 & {\bf0.2032} & 6.331 \\
-- & \checkmark & 3 & -- & 30.32 & 0.6374 & 0.2064 & 6.144 \\
-- & \checkmark & 1 & \checkmark & {\bf30.36} & {\bf0.6401} & 0.2236 & {\bf5.338} \\

\bottomrule
\end{tabular}
}

\label{tab:ablation}
\end{table}
\subsubsection{Video Denoising}
Video denoising is typically more straightforward for supervised methods, as it does not require generating new content or inferring missing details. 
Despite this, as demonstrated in Table~\ref{tab:dn}, our method outperforms SoTA supervised video denoising method, Shift-Net~\citep{li2023simple} in terms of WE and FSim and exhibits strong robustness, maintaining consistent performance even as noise levels increase. As shown in Fig.~\ref{fig:dn}, denoising results on the SPMC dataset highlight the limitations of Inf-DiT and DiffIR2VR-Zero, which fail to recover fine details such as textures on sculptures and numbers on license plates. In contrast, our approach achieves a superior balance between restoring intricate details and ensuring temporal consistency across frames.
\textit{Additional results on other restoration tasks are in Supp. material.}



\subsection{Ablation Studies}
We present an ablation study to demonstrate the effectiveness of the three key components in our proposed video restoration framework as shown in Table~\ref{tab:ablation}. Using Inf-DiT as the baseline configuration (denoted as all (--)), optical flow warping alone provides modest improvements but shows persistently low SSIM values due to Vid4's small dataset size leading to erroneous frame generation that warping cannot rectify. The inclusion of TAttn yields a substantial performance improvement of 1.81 dB in PSNR and a significant reduction in warping error, underscoring its role in capturing critical temporal dependencies. Adding STNC with 1 neighbor per frame provides substantial improvements across all metrics, while increasing to 3 neighbors yields only marginal performance gains at significantly higher computational cost, indicating that single neighbor selection offers the optimal efficiency-performance trade-off. Finally, the integration of the flow-guided sampler leads to notable improvements in performance, highlighting its importance in achieving precise motion alignment. It is evident that the complete model, which combines all three components, delivers the highest performance across most metrics, validating the complementary contributions of each module (\textit{Additional ablations in Supp.}).

    

\section{Conclusion}
\label{sec:conclusion}
In this paper, we proposed DiTVR, a zero-shot video restoration framework leveraging a pre-trained Diffusion Transformer in pixel space. By integrating a spatiotemporal neighbor cache mechanism, trajectory-aware attention, and a flow-guided diffusion sampler, DiTVR achieves enhanced global and local details, and robust temporal consistency. Extensive experiments demonstrate that DiTVR outperforms existing methods, excelling in temporal coherence and high-fidelity restoration without any task-specific training.

\bibliography{aaai2026}

\end{document}